\documentclass[conference]{IEEEtran}
\IEEEoverridecommandlockouts
\usepackage{cite}
\usepackage{amsmath,amssymb,amsfonts}
\usepackage{algorithm}
\usepackage{algpseudocode}
\usepackage{graphicx}
\usepackage{textcomp}
\usepackage{xcolor}

\usepackage[colorinlistoftodos]{todonotes}
\usepackage{acronym}
\usepackage{adjustbox}
\usepackage{subfigure}
\usepackage{color}
\usepackage{varioref} 
\usepackage[hidelinks]{hyperref}
\usepackage{siunitx}  
\usepackage{tabularx}
\usepackage{stfloats}
\usepackage{float}
\usepackage{placeins}
\usepackage{balance}

\DeclareSIUnit{\kn}{knots}

\newacro{SSS}{Side Scan Sonar}
\newacro{SONAR}{SOund NAvigation and Ranging}
\newacro{AWBs}{Artificial Water Bodies}
\newacro{USV}{Unmanned Surface Vehicle}
\newacro{ROV}{Remotely Operated Vehicle}
\newacro{NDVI}{Normalized Difference Vegetation Index}
\newacro{SAV}{Submerged Aquatic Vegetation}
\newacro{NDWI}{Normalized Differential Water Index}
\newacro{FAI}{Floating Algae Index}
\newacro{APA}{Aquatic Plants and Algae}
\newacro{AOI}{areas of interest}
\newacro{OSM}{OpenStreetMap}
\newacro{CNN}{Convolutional Neural Network}

\def\BibTeX{{\rm B\kern-.05em{\sc i\kern-.025em b}\kern-.08em
    T\kern-.1667em\lower.7ex\hbox{E}\kern-.125emX}}

\begin{document}


\title{Towards Robotic Lake Maintenance: Integrating SONAR and Satellite Data to Assist Human Operators%
\thanks{\copyright 2026 IEEE. Personal use of this material is permitted. Permission from IEEE must be obtained for all other uses, in any current or future media, including reprinting/republishing this material for advertising or promotional purposes, creating new collective works, for resale or redistribution to servers or lists, or reuse of any copyrighted component of this work in other works.}
}



\author{\IEEEauthorblockN{Ahmed H. Elsayed\textsuperscript{*}, Christoph Manss, Tarek A. El-Mihoub, Andrej Lejman, Frederic Stahl}
\IEEEauthorblockA{\textit{Marine Perception Department} \\
\textit{German Research Center for Artificial Intelligence (DFKI), Oldenburg, Germany}\\
\textsuperscript{*}Corresponding author: ahmed.elsayed@dfki.de}
}

\maketitle


\begin{abstract}
\ac{AWBs} are human-made systems that require continuous monitoring due to their artificial biological processes. These systems demand regular maintenance to manage their ecosystems effectively. As a result of these artificial conditions, underwater vegetation can grow rapidly and must be harvested to preserve the ecological balance. This paper proposes a two-step approach to support targeted weed harvesting for the maintenance of artificial lakes. The first step is the initial detection of \ac{SAV}, also referred to in this paper as areas of interest, is performed using satellite-derived indices, specifically the \ac{APA} index, which highlights submerged vegetation in water bodies. Subsequently, an \ac{USV} equipped with multibeam \ac{SONAR} performs high-resolution bathymetric mapping to locate and quantify aquatic vegetation precisely.
This two-stage approach offers an effective human-robot collaboration, where satellite data guides the \ac{USV} missions and boat skippers leverage detailed SONAR maps for targeted harvesting. This setup narrows the search space and reduces manual workload from human operators, making the harvesting process less labour-intensive for operators. Preliminary results demonstrate the feasibility of integrating satellite imagery and underwater acoustic sensing to improve vegetation management in artificial lakes.

\end{abstract}


\begin{IEEEkeywords}
Unmanned Surface Vehicles, Aquatic Vegetation, Remote Sensing, Multibeam SONAR, Human-Robot Collaboration 
\end{IEEEkeywords}

\section{Introduction}

\acf{AWBs} are created by humans for different reasons, such as water retention in dam construction, urban development, rainwater storage, or leisure activities. Unlike natural lakes, \ac{AWBs} lack an established ecosystem, making them prone to environmental issues like ecological degradation and the spread of diseases if not properly monitored and maintained \cite{hunter1982man,Charudattan2001,Clayton1996}. Managing \ac{AWBs}, including artificial lakes, presents unique challenges due to their reliance on continuous human intervention, given the unnatural balance of their ecosystems and the potential for disturbances in biological processes \cite{cantonati2020characteristics}.
The Maschsee, a lake in Hannover, Germany, is a representative \ac{AWBs} facing persistent challenges due to the rapid weed growth, which poses a significant problem that requires continuous harvesting~\cite{haz2025}. Effective monitoring could reduce human labour and improve the efficiency of maintaining the lake's ecosystem. The proliferation of weeds disrupts aquatic ecosystems, negatively impacting the lake's biological life, affecting fish and other aquatic species, and impairing leisure activities such as kayaking \cite{Lange2000}. However, weed accumulations can obstruct the propellers of small boats, posing risks to humans and equipment. 


Mechanical mowing is a commonly used method for managing excessive aquatic vegetation in lakes, particularly in areas with high recreational or ecological value. It helps improve aesthetics, usability, and ecological balance by removing biomass that can cause oxygen fluctuations and hinder biodiversity \cite{matheson2015purpose,Lange2000}. Additionally, mowing contributes to nutrient removal, which can help prevent the buildup of nutrients that lead to poor water quality. However, if done excessively, it may disrupt nutrient cycling and affect the overall water quality. Mechanical harvesters typically operate at depths of 2–3 meters and have constraints in speed and storage capacity, making targeted and efficient harvesting essential. These limitations highlight the importance of accurate vegetation mapping to optimise mowing operations and reduce unnecessary effort.

Efficient monitoring of such lakes is crucial for sustainable maintenance and energy conservation. As pointed out in \cite{Inacio2023}, lake ecosystems have to be mapped to enable municipalities to do effective maintenance. 
In this work, we propose a heterogeneous monitoring system combining satellite imagery, \ac{SONAR} equipped \ac{USV}, and human operators. The proposed system enables the autonomous collection of data to support real-time decision-making by the lake maintenance crews.

We use satellite imagery to provide a high-level overview and a low-effort method to detect \acp{SAV} \cite{rowan2021review}. 
For example, in \cite{Rodriguez-Garlito2023}, neural networks are used to detect invasive plants in satellite imagery using spectral indices. In particular, multiple indices were computed and utilised as input to a \ac{CNN}, which is then trained in a supervised approach.
However, for precise \ac{SAV} mapping, acoustic sensing such as \ac{SONAR} is more effective than optical sensors, particularly in turbid or deep waters \cite{mutlu2023acoustic, komatsu2003use}. While previous research has explored both \ac{SONAR} and optical cameras for \ac{SAV} monitoring \cite{gerlo2023seaweed}, \ac{SONAR} and backscatter strength data outperform optical methods in \ac{SAV} \cite{mutlu2023acoustic, komatsu2003use,greene2018side}. \ac{SONAR} is particularly advantageous due to its ability to scan larger areas at greater depths compared with optical cameras, though it requires specialised expertise for operation and data interpretation. By leveraging \ac{SONAR}, we can achieve more efficient monitoring of \ac{SAV}.

Although satellite and acoustic sensing methods each support monitoring \ac{SAV} independently \cite{huber2021novel,bennett2020using}, their integration remains relatively unexplored.
To address these challenges, we propose a heterogeneous mission setup that combines satellite imagery with a \ac{USV} equipped with \ac{SONAR}. This system supports real-time collaboration with human operators by streaming \ac{SONAR} data to operators during \ac{USV} missions, enabling targeted \ac{SAV} harvesting and efficient, scalable monitoring of artificial lakes.

The main contributions of this paper are:
\begin{itemize}
    \item Low-resolution \ac{SAV} detection using satellite imagery.
    \item High-resolution \ac{SAV} detection using a \ac{USV} equipped with a multibeam \ac{SONAR}.
    \item A heterogeneous mission setup that takes advantage of satellite and \ac{SONAR} sensing to support targeted \ac{SAV} harvesting in artificial lakes.
\end{itemize}

The rest of the paper is organised as follows: Section \ref{sec:mission_setup} shows the mission setup and the methodology, Section \ref{sec:results} details the experimental results and our findings, and Section \ref{sec:conc} discusses the conclusion and potential future directions.


\section{Methodology \& Mission Setup} \label{sec:mission_setup}
\begin{figure}[ht]
  \centering
  \includegraphics[width=\columnwidth]{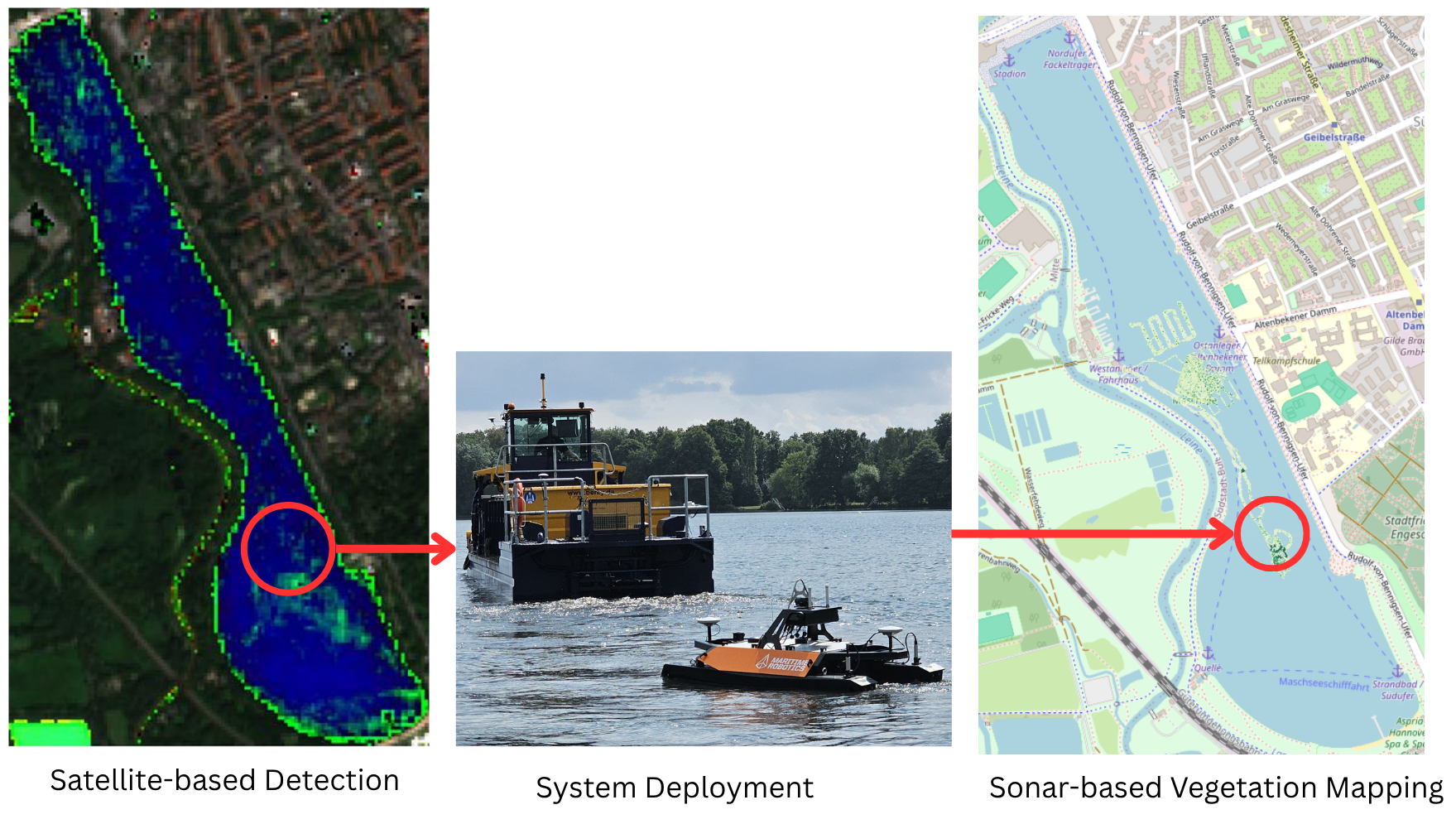}
  \caption{Mission setup. \textbf{1-Satellite detection:} potential \ac{SAV} clusters marked in green. \textbf{2-On‑site survey:} \ac{USV} and harvester deployed. \textbf{3-Vegetation map:} SONAR‑derived weed height overlaid on Maschsee Lake map.}
  \label{fig:mission_setup}
\end{figure}

The mission supports the weed harvesting process in Maschsee Lake by adopting a heterogeneous system combining human operators, satellite imagery, and an \ac{USV}. The aim is to detect, map, and localise \ac{SAV} for removal, enabling targeted harvesting. Figure~\ref{fig:mission_setup} shows the workflow, starting with the initial detection from the satellite, then the on-site survey with the \ac{USV} and the harvester, to vegetation mapping using \ac{SONAR}. 

The \ac{USV} used in this study is a fully autonomous robotic surface vehicle equipped with a multibeam \ac{SONAR} and onboard GPS for precise navigation.\footnote{https://www.maritimerobotics.com/otter, accessed on 28-07-2025} It operates at speed up to \SI{6}{\kn} and supports real-time data transmission to shore-based operators. The vehicle's autonomy enables it to follow predefined survey paths within the target areas identified from the satellite data, allowing for highly efficient high-resolution mapping of \ac{SAV}.

\subsection{Satellite Imagery For Vegetation Detection}

\Ac{AOI} for \ac{SAV} can be detected using satellite imagery from the Sentinel-2 mission. Sentinel-2 operates on a 5-day revisit cycle over Earth's landmass. It provides 10 optical bands suitable for observing natural processes and for developing machine learning and AI-based algorithms. The spatial resolution of the relevant Sentinel-2 bands used for aquatic vegetation detection is 10 meters per pixel. These spectral bands are typically combined into indices that highlight specific biophysical phenomena. Many such indices are available via Sentinel Hub.\footnote{\url{https://www.sentinel-hub.com/}, accessed on 16-07-2025}

In this paper, we utilise the \acf{APA} index from \cite{peliova-anna-aquatic-nodate} to detect \ac{SAV} using Sentinel-2 data. The \ac{APA} index is a composite index in the RGB format, where the red (R), green (G), and blue (B) channels represent the \ac{FAI}, \ac{SAV} index, and the \ac{NDWI}, respectively. Each component is scaled to enhance visual contrast and is displayed only when a pixel is classified as water; otherwise, the true colour image is shown. In this paper, only the \ac{SAV} index is of interest. Thus, the second channel of the \ac{APA} index is calculated as,
\begin{equation}
    G = \frac{\text{RED\_EDGE} - \text{RED}}{\text{RED\_EDGE} + \text{RED}},
    \label{eq:apa_green_channel}
\end{equation}
\noindent where RED\_EDGE and RED refer to optical channels of the Sentinel-2 satellite. The values of the green channel represent a normalised difference index. The index ranges between $-1$ and $1$, but we scale it into integer values ranging from $0$ to $255$. 
Higher values, $G=255$, indicate a greater density of submerged or floating aquatic plants. Lower values $G=0$ correspond to open water with little vegetation. However, values close to $G=255$ are most likely misinterpreted as regular vegetation, especially at the shore of the lake. Thus, the resulting index is then processed further to detect aquatic vegetation. 

First, the image of the index gets cropped according to the boundaries of the lake, which are taken from OpenStreetMap.\footnote{https://www.openstreetmap.org/, accessed on 28-07-2025}
Then, vegetation within the lake boundaries is visible in green colour. To distinguish between vegetation on the shore and the water, a k-means clustering on all possible values in the index is applied, with $k_i=5$ classes ranging from  no, low, medium, and high aquatic plant density to on-shore vegetation.
Next, relevant aquatic plant densities are considered to map them into relevant regions. Here, we consider medium and high aquatic plant densities, and use the k-means algorithm again with $k_a=15$ to cluster the intensity map into 15 \acp{AOI}.
Each \ac{AOI} can then be harvested with the weed harvester.
This process is summarised in Algorithm~\ref{alg:aoi_detection}.
\begin{algorithm}
    \caption{Detection of \aclp{AOI} for weed harvesting from the APA index}\label{alg:aoi_detection}
    \begin{algorithmic}[1]
        \Require \text{location}, \text{date}
        \State $\text{boundaries} \gets \text{OSM}(\text{location})$
        \State $\text{APA\_index} \gets \text{Sentinel API}(\text{location, date})$
        \State $G \gets \text{Crop}(\text{boundaries, APA\_index})$
        \State $\text{intensity\_cluster\_centroids} \gets \text{k-means}(G, k_i=5)$
        \State $\text{areas\_of\_interest} \gets \text{k-means}(\text{intensity\_cluster\_centroids},$\newline$ k_a=15)$
        \State \Return $\text{areas\_of\_interest}$
    \end{algorithmic}
\end{algorithm}


\subsection{Underwater SONAR Mapping}

Following the initial satellite detection, underwater mapping via \ac{SONAR} enables targeted and efficient weed harvesting by focusing only on \acp{AOI}, reducing the overall survey time for the \ac{USV}.

For the survey, the Norbit iWBMS multibeam SONAR is operated at a mean frequency of \SI{400}{\kilo\hertz}, using an \SI{80}{\kilo\hertz} chirp to form 256 beams with \SI{0.9}{\degree} resolution.\footnote{https://norbit.com/subsea/products, accessed on 28-07-2025} The system covers a \SI{150}{\degree} swath at an average survey speed of \SI{3}{\kn} for dense bathymetric mapping. The upper and lower gates were initially set to \SI{1.0}{\metre} and \SI{5.0}{\metre}, respectively, based on general assumptions of the lake depth to reduce surface and bottom noise.

 The \ac{USV} performs a pre-harvest scan of defined areas using its multibeam \ac{SONAR} to map underwater weed clusters. Once the area is fully surveyed, the weed harvester boat follows the \ac{USV}'s path to collect the weeds. A post-harvest scan by the \ac{USV} enables comparison of the lake’s condition before and after harvesting.

The \ac{SONAR} data were recorded using the BeamWorX sub-software \textit{NavAQ}, then processed and cleaned using \textit{AutoClean}.\footnote{https://www.beamworx.com/, accessed on 04-08-2025} The resulting weed height map is generated as a span layer, representing the difference between the shallowest and deepest returns, and exported as a GeoTIFF at \SI{0.1}{\metre} resolution.

\begin{figure}[htp]
 \centering
 \includegraphics[width=0.8\columnwidth]{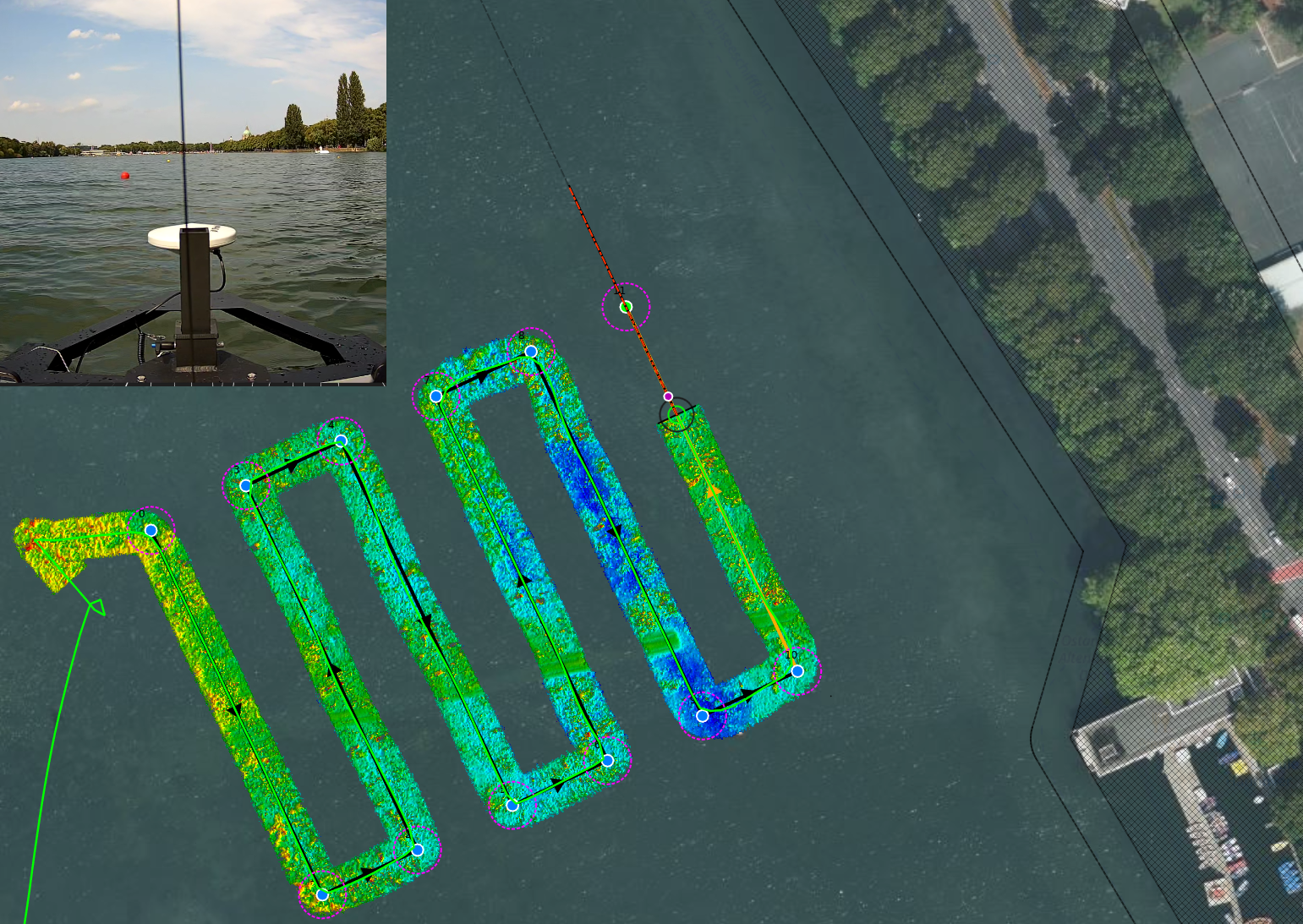}
 \caption{Bathymetric map of Maschsee Lake with overlaid RGB imagery from the \ac{USV}'s onboard camera.}
 \label{fig:vcs_rgb}
\end{figure}

To support real-time decision-making during harvesting, the \ac{USV} continuously transmitted georeferenced bathymetric and backscatter data to a surface interface. This interface visualises the underwater topography of the lake, enabling operators to plan efficient harvesting routes based on depth variations and potential obstacles. An example of this integrated visualisation is shown in Figure~\ref{fig:vcs_rgb}.

To generate a vegetation height map, BeamWorx Autoclean software is used to process the multibeam \ac{SONAR} data. The workflow involves importing the raw data, applying sound velocity corrections, and isolating the vegetation layer using upper and lower gates. Then, the software computes the span between the shallowest and deepest returns. It is worth noting that this method may also capture submerged structures or objects, which can be misinterpreted as \ac{SAV}. To mitigate this, backscatter intensity is used as an additional indicator to identify hard surfaces that might pose a risk to the weed harvester. This step is performed manually through visual inspection to mark areas of avoidance, but no automated removal is applied to the exported vegetation map. This process is summarised in Algorithm~\ref{alg:veg_map}.

\begin{algorithm}
    \caption{Generation of vegetation height map using BeamWorX}\label{alg:veg_map}
    \begin{algorithmic}[1]
        \Require \text{Survey area}, \text{SONAR configuration}
        \State \text{Import raw multibeam SONAR data into BeamWorX}
        \State \text{Apply sound velocity profile for depth correction}
        \State \text{Set upper and lower gates to isolate vegetation layer}
        \State \text{Compute span: } $\text{span} = d_{\text{bottom}} - d_{\text{top}}$
        \State \text{Inspect backscatter intensity to identify hard objects}
        \State \text{Note areas to avoid (manual only, no masking applied)}
        \State \text{Export weed height map as GeoTIFF}
    \end{algorithmic}
\end{algorithm}

\section{Experimental Results \& Findings} \label{sec:results}
This section shows the results of the proposed two-stage approach, detecting and validating \ac{SAV} in Maschsee Lake.
The approach combines satellite-based remote sensing with in-situ \ac{SONAR} validation.

\begin{figure*}[htp]
    \centering
    \includegraphics[width=\linewidth]{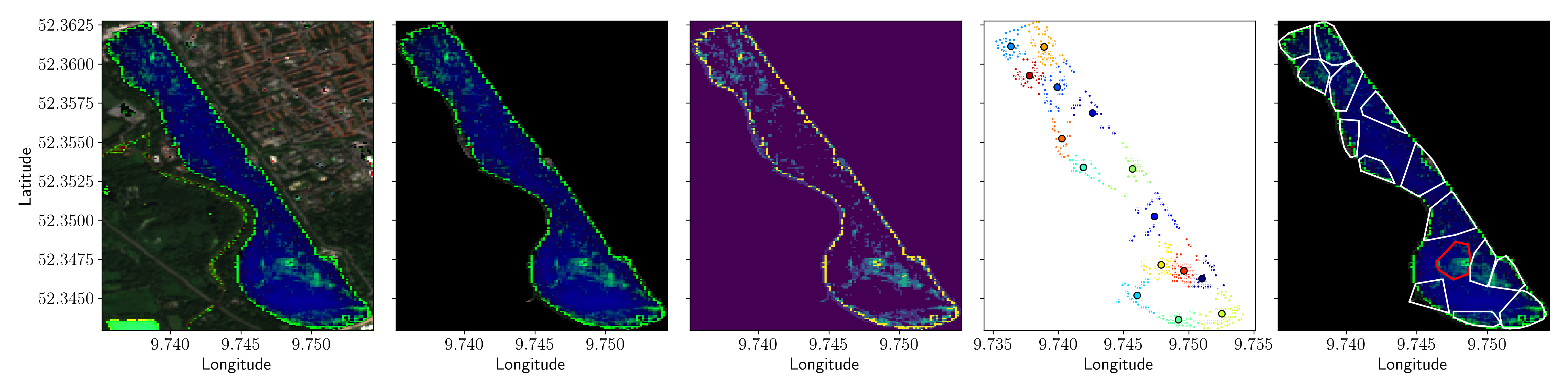}
    \caption{Detection of relevant areas with aquatic plants. From left to right, this figure depicts the \ac{APA} index for the 6\textsuperscript{th} of August 2024, the cropping to the lake boundaries, the clustering according to vegetation intensity, the clustering according to location, and the resulting areas with relevant submerged vegetation.}
    \label{fig:aoi_detection}
\end{figure*}

\subsection{Stage 1: Satellite}
We present the \ac{AOI} detection for the 6\textsuperscript{th} of August 2024, which is the closest possible date of a Sentinel-2 fly over prior to the \ac{SONAR} measurements conducted on the 19\textsuperscript{th} and 20\textsuperscript{th} of August 2024. While Sentinel-2 revisits the area every five days, the images captured after the 6\textsuperscript{th} of August 2024 were occluded by cloud cover.

Each intermediate result of the steps in Algorithm~\ref{alg:aoi_detection} is visualised in Figure~\ref{fig:aoi_detection}.
From left to right, Figure~\ref{fig:aoi_detection} shows the computed \ac{APA} index, the cropped \ac{APA} index, the clustered intensities, the clustered regions, and the resulting areas of interest.
Several \acp{AOI} were identified through this process, representing regions with potential \ac{SAV}. Among these, the area around the centroid at $(9.7475, 52.346)$ was of particular interest, as this represents a high density of \ac{SAV} and is also investigated further with the \ac{SONAR}.

Thus, the identified \acp{AOI} serve as a basis for guiding in-situ measurements. In particular, the aforementioned region was used to compute remote sensing results from the satellite with direct underwater observations with the multibeam \ac{SONAR}.
While satellite data provide wide spatial coverage and early indicators of vegetation presence, it remains limited by temporal resolution, cloud cover, and water turbidity. These factors justify the need for complementary in-situ measurements when precise or real-time vegetation assessment is required.

\begin{figure}[thpb]
  \centering
  \includegraphics[width=0.9\columnwidth]{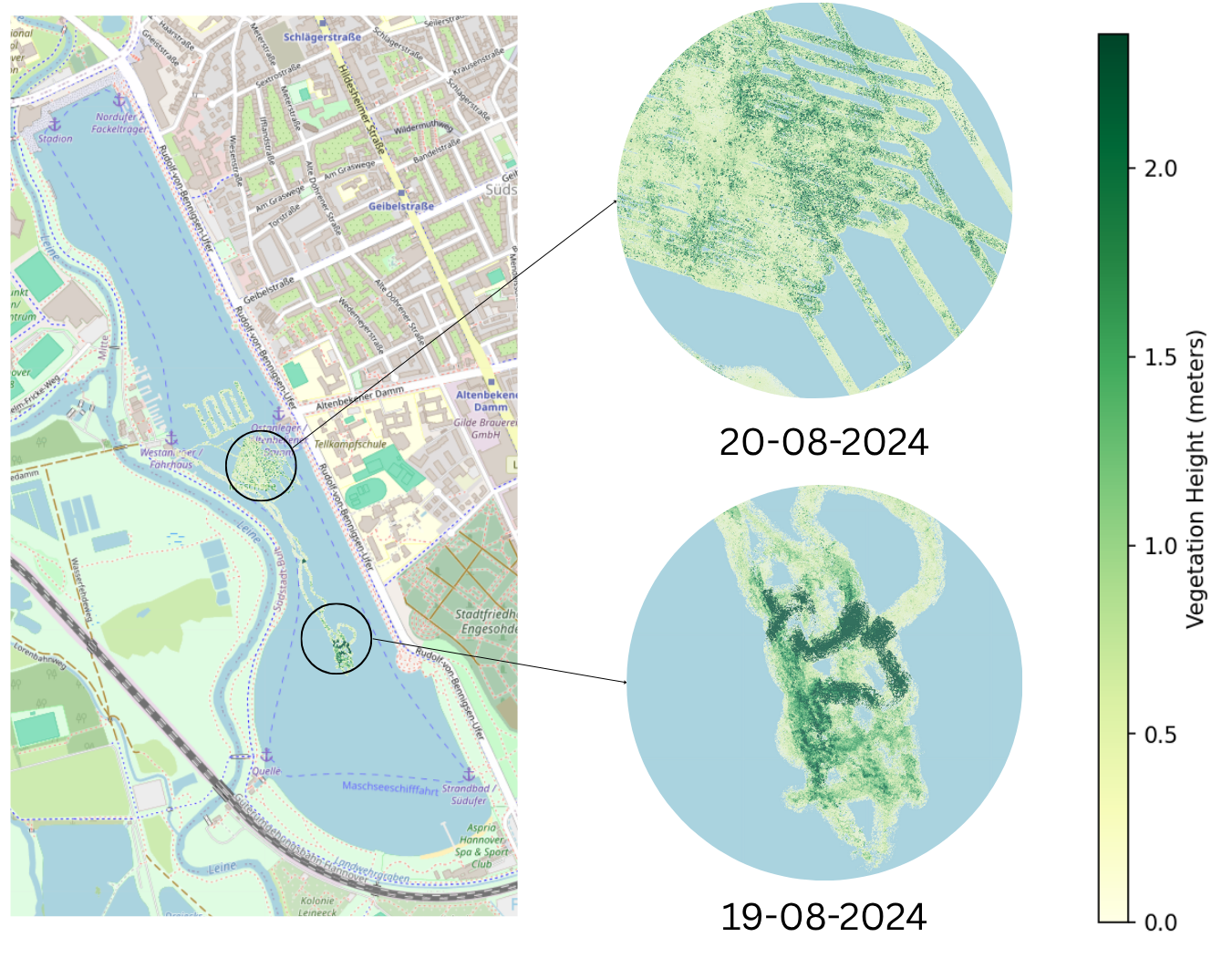}
  \caption{Vegetation map derived from SONAR bathymetry using elevation difference, based on surveys conducted on the 19\textsuperscript{th} and 20\textsuperscript{th} of August 2024.}
  \label{fig:vegetation_map_overall}
\end{figure}

\subsection{Stage 2: SONAR-based Mapping results}

The multibeam \ac{SONAR} survey successfully maps the \ac{SAV} in Maschsee Lake with high spatial resolution and produces vegetation maps on 19\textsuperscript{th} and 20\textsuperscript{th} of August 2024. These vegetation maps enable boat skippers to visualise weed distribution, allowing for targeted harvesting operations. Data were recorded using the BeamWorx sub-software NavAQ, then processed and cleaned using AutoClean.

Using SONAR data, weed areas can be identified based on backscatter intensity and bathymetric elevation. Due to the hard interpretation of the backscatter data and the relative simplicity of the bathymetric data, weed areas are identified using bathymetric elevation. In this method, we use sonar-derived raster data from minimum and maximum elevation values taken from the same scan over a specific area. By subtracting the minimum from the maximum elevation, we get local height differences that most likely correspond to weeds. The resulting multibeam bathymetric data revealed dense weed clusters across several regions, as shown in Figure~\ref{fig:vegetation_map_overall}. This high-resolution mapping supplements the coarser satellite-derived \ac{AOI} by offering precise information which is hard to obtain with remote sensing.

\begin{figure}[thpb]
  \centering
  \includegraphics[width=\columnwidth]{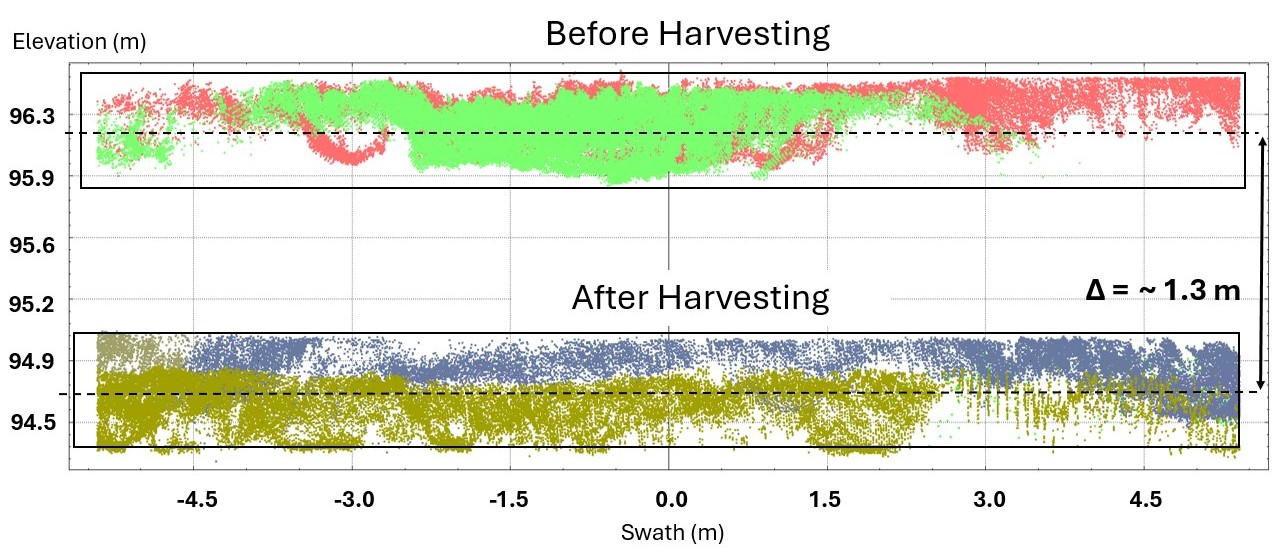}
  \caption{2D view of \ac{SONAR} scan from an inspection area, showing the height difference in weed distribution detected by multibeam \ac{SONAR} \textbf{[Top]} before harvesting and \textbf{[Bottom]} after harvesting, showing an average difference of \SI{1.3}{\meter}. Each color cluster represents a single scan from the SONAR.}
  \label{fig:before_after_mowing}
\end{figure}

To evaluate the effectiveness of weed removal and assess bathymetric data quality, a designated area is scanned before and after harvesting. The average height difference between the two scans was found to be \SI{1.3}{\meter}, confirming the removal of dense vegetation. Figure~\ref{fig:before_after_mowing} shows the elevation maps before and after harvesting, illustrating the change in weed distribution. This quantitative result not only validates the harvesting process but also provides a baseline for estimating biomass removal and optimising future harvesting strategies.

\begin{figure}[thpb]
    \centering
    \subfigure[\label{fig:ladder_maschsee_gt}]{\adjincludegraphics[width=.49\columnwidth, trim={0 0 {.5\width} 0}, clip=true]{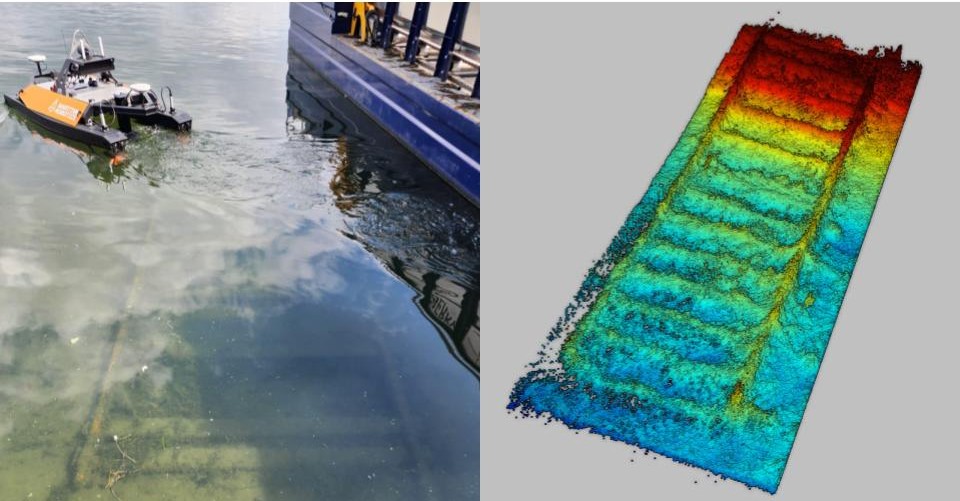}}
    \subfigure[\label{fig:ladder_beamworx}]{\adjincludegraphics[width=.49\columnwidth, trim={{.5\width} 0 0 0}, clip=true]{images/ladder_camera_sonar.jpg}}
    \subfigure[\label{fig:ladder_backscatter}]{\adjincludegraphics[width=.8\columnwidth, trim={0 0 0 0}, clip=true]{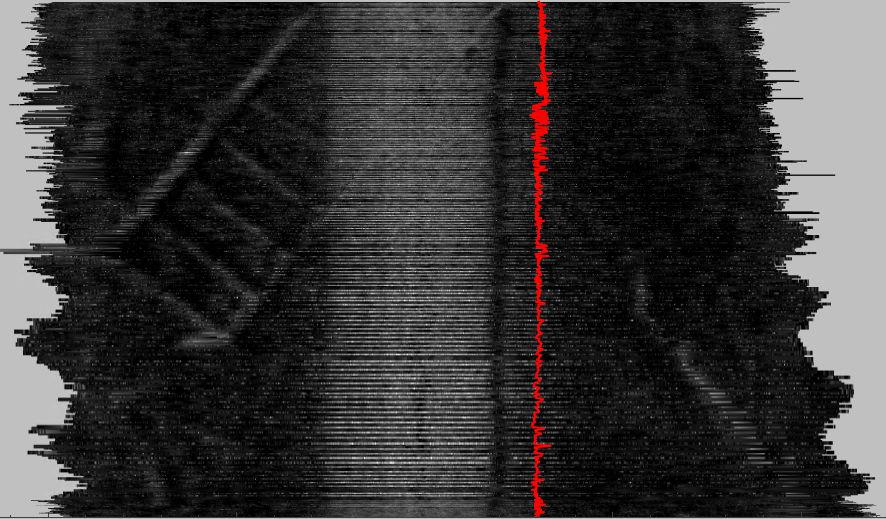}}
    \caption{Submerged boat launch rail [\textbf{a}] Camera image above water [\textbf{b}] Multibeam \ac{SONAR} 3D point cloud data [\textbf{c}] \ac{SONAR} Backscatter data, where the red line indicates the trajectory of the \ac{USV} during data acquisition.}
    \label{fig:ladder_maschsee}
\end{figure}

In addition to bathymetry, backscatter data are recorded to support the classification of underwater targets. The backscatter data allowed differentiation between dense vegetation, lakebed, and submerged structures. For instance, a submerged boat launch rail was detected and confirmed through a combination of \ac{SONAR} point cloud, backscatter intensity, and camera imagery as shown in Figure~\ref{fig:ladder_maschsee}.

\begin{figure}[htpb]
  \centering
  \includegraphics[width=0.8\columnwidth]{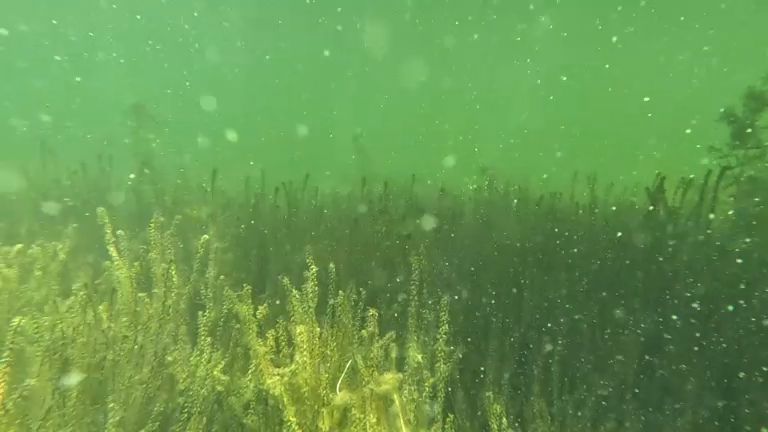}
  \caption{Underwater image captured with a GoPro camera showing \ac{SAV} in Maschsee Lake, used for qualitative validation of SONAR-based weed detection.}
  \label{fig:gopro_maschsee}
\end{figure}

Underwater footage from a GoPro camera was also used to qualitatively validate the \ac{SONAR} results in the same area. Figure~\ref{fig:gopro_maschsee} confirms the presence of vegetation detected by the \ac{USV}, supporting the interpretation of bathymetric elevation differences as \ac{SAV}.

Overall, the SONAR-based mapping provided depth and reliable spatial data, confirming both the presence and removal of \ac{SAV}. It further complements satellite-based detection by overcoming limitations due to water turbidity, temporal resolution, rainfall, etc.

\section{Discussion}\label{sec:conc}

This work introduced a mission setup for detecting \ac{SAV} to support targeted weed harvesting in artificial lakes. Low-resolution satellite imagery was first used to identify potential \ac{AOI}. Some of these regions were then surveyed by an \ac{USV} equipped with a multibeam \ac{SONAR}, enabling high-resolution in-situ measurements. This two-stage approach allows for efficient pre-selection of survey sites via remote sensing, followed by detailed underwater mapping. Bathymetric data was used to detect \ac{SAV}, while backscatter strength helps to distinguish potential underwater objects from \ac{SAV}s to avoid them while doing the lake maintenance. This approach demonstrates how the \ac{USV} as a robotic platform can support human operators to help and facilitate the maintenance and reduce the labour intensity by providing detailed maps to the boat skippers for the harvesting process.

Future work will focus on several key areas to enhance the weed harvesting process through several key aspects. A primary objective is to provide a detailed analysis of bathymetry and backscatter data from the \ac{SONAR} to correlate it with \ac{SAV} volume and distribution across the lake. This analysis aims to refine our understanding of how \ac{SONAR} data reflects weed density, which helps in planning the capacity of the weed harvester.

To validate the \ac{SONAR}-based detection, a \ac{ROV} equipped with an underwater camera and an acoustic positioning system will be deployed. This will allow precise location tracking of the ROV, providing ground truth for weed distribution.

Currently, weed detection relies on manual visual inspection by human operators, which is time-consuming. As a potential direction, Artificial Intelligence techniques could be explored to automate this process. Image segmentation methods, such as those demonstrated in \cite{garone2023seabed}, may enhance detection efficiency and accuracy.

This study was conducted on a single artificial lake and focused on one detection-harvest cycle. As such, the repeatability and generalisability of the results to other water bodies or operational conditions remain to be validated in future work.

Another possible area for future development is designing and implementing a path-planning algorithm to optimise the weed harvesting workflow. This algorithm will calculate the most efficient route for the boat, assisting the skipper in navigating through the weed clusters and harvesting them efficiently. It will also ensure that the boat's capacity (\SI{15}{\metre\cubed}) is not exceeded, allowing the boat to reach the unloading station at the optimal time. Additionally, integrating dynamic updates based on real-time data from the \ac{USV} will enable the algorithm to adjust the route in response to newly detected weed clusters during the mission. Developing this functionality will significantly enhance the efficiency and adaptability of the harvesting process.

\section*{Acknowledgment}

This work is done within the HAI-x project, which is funded by the BMBF (Funding number: 01IW23003).

\bibliographystyle{ieeetr}
\bibliography{ref,cmanss}
\balance

\end{document}